\documentclass{article}

\usepackage[nonatbib,final]{neurips_2021}
\usepackage[utf8]{inputenc} % allow utf-8 input
\usepackage[T1]{fontenc}    % use 8-bit T1 fonts
\usepackage{hyperref}       % hyperlinks
\usepackage{url}            % simple URL typesetting
\usepackage{booktabs}       % professional-quality tables
\usepackage{amsfonts}       % blackboard math symbols
\usepackage{nicefrac}       % compact symbols for 1/2, etc.
\usepackage{microtype}      % microtypography
\usepackage{xcolor}         % colors
\usepackage{graphicx}
\usepackage{enumitem}

\usepackage[font=small,labelfont=bf]{caption}

\usepackage{scrextend}

\title{On Efficient Uncertainty Estimation for Resource-Constrained Mobile Applications}

\author{%
	Johanna Rock\thanks{Work performed while the author was at Arm ML Research Lab} \\
	SPSC Lab\\
	Graz, University of Technology\\
	\texttt{johanna.rock@tugraz.at} \\
	\And
	Tiago Azevedo \\
	Arm ML Research Lab \\
	\texttt{tiago.azevedo@arm.com} \\
	\And
	René de Jong \\
	Arm ML Research Lab \\
	\texttt{rene.dejong@arm.com} \\
	\And
	Daniel Ruiz-Muñoz \\
	Arm ML Research Lab \\
	\texttt{daniel.ruizmunoz@arm.com} \\
	\And
	Partha Maji \\
	Arm ML Research Lab \\
	\texttt{partha.maji@arm.com} \\
}

\begin{document}
	
	\maketitle
	
	\setcounter{footnote}{0}
	
	\begin{abstract}
	
	Deep neural networks have shown great success in prediction quality while reliable and robust uncertainty estimation remains a challenge. Predictive uncertainty supplements model predictions and enables improved functionality of downstream tasks including embedded and mobile applications, such as virtual reality, augmented reality, sensor fusion, and perception. These applications often require a compromise in complexity to obtain uncertainty estimates due to very limited memory and compute resources. We tackle this problem by building upon Monte Carlo Dropout (MCDO) models using the Axolotl framework; specifically, we diversify sampled subnetworks, leverage dropout patterns, and use a branching technique to improve predictive performance while maintaining fast computations. We conduct experiments on (1) a multi-class classification task using the CIFAR10 dataset, and (2) a more complex human body segmentation task. Our results show the effectiveness of our approach by reaching close to Deep Ensemble prediction quality and uncertainty estimation, while still achieving faster inference on resource-limited mobile platforms.

	\end{abstract}
	
	% keywords:
	% predictive uncertainty, monte carlo dropout, mcdo, contrastive dropout, branched computation, resource-efficient, fast inference
	
	% TL;DR
	% Introduces an approach on efficient uncertainty estimation for resource-constrained mobile applications by extending MCDO with spatial convolutions, contrastive dropout and batched computation of branched networks.
	
	\section{Introduction}
	Uncertainty estimation and out-of-distribution robustness are vital aspects in modern deep learning. Predictive uncertainty supplements model predictions and enables improved functionality of downstream tasks including various resource-constrained embedded and mobile applications. Popular examples are virtual reality (VR), augmented reality (AR), sensor fusion, perception, and health applications including fitness indicators, arrhythmia detection, and skin lesion detection. Robust and reliable predictions with uncertainty estimates are increasingly important when operating on noisy in-the-wild data from sensory inputs. A large variety of deep learning architectures have been applied to various tasks with great success in terms of prediction quality, however, producing reliable and robust uncertainty without additional computational and memory overhead remains a challenge~\cite{ovadia2019can}. This issue is further aggravated due to the limited computational and memory budget available in typical battery-powered mobile devices.
	
	There exist many approaches towards uncertainty estimation, however, many of them are complex to train, lack good predictive performance, or are very resource-intensive~\cite{vadera2020ursabench}. One of the best performing approaches is Deep Ensemble~\cite{lakshminarayanan2016simple}, where multiple models are trained independently and used at inference, thus making it not viable on mobile platforms with low-latency requirements. A more resource-efficient approach is Monte Carlo Dropout (MCDO)~\cite{gal2016dropout}, where dropout is used during training and inference on a single model, yielding an approximate distribution of predictions due to dropout-based sampling. MCDO has been shown to produce less diverse predictions than Deep Ensemble~\cite{fort2019deep} and therefore achieve poor predictive performance in terms of prediction quality as well as uncertainty estimates. There are multiple other interesting approaches~\cite{huang2017snapshot,havasi2020training} for efficient uncertainty estimation, however, in this work we focus on improving MCDO due to its simplicity and low-memory footprint.
	
\begin{figure}[th]
		\centering
		\includegraphics[width=\textwidth]{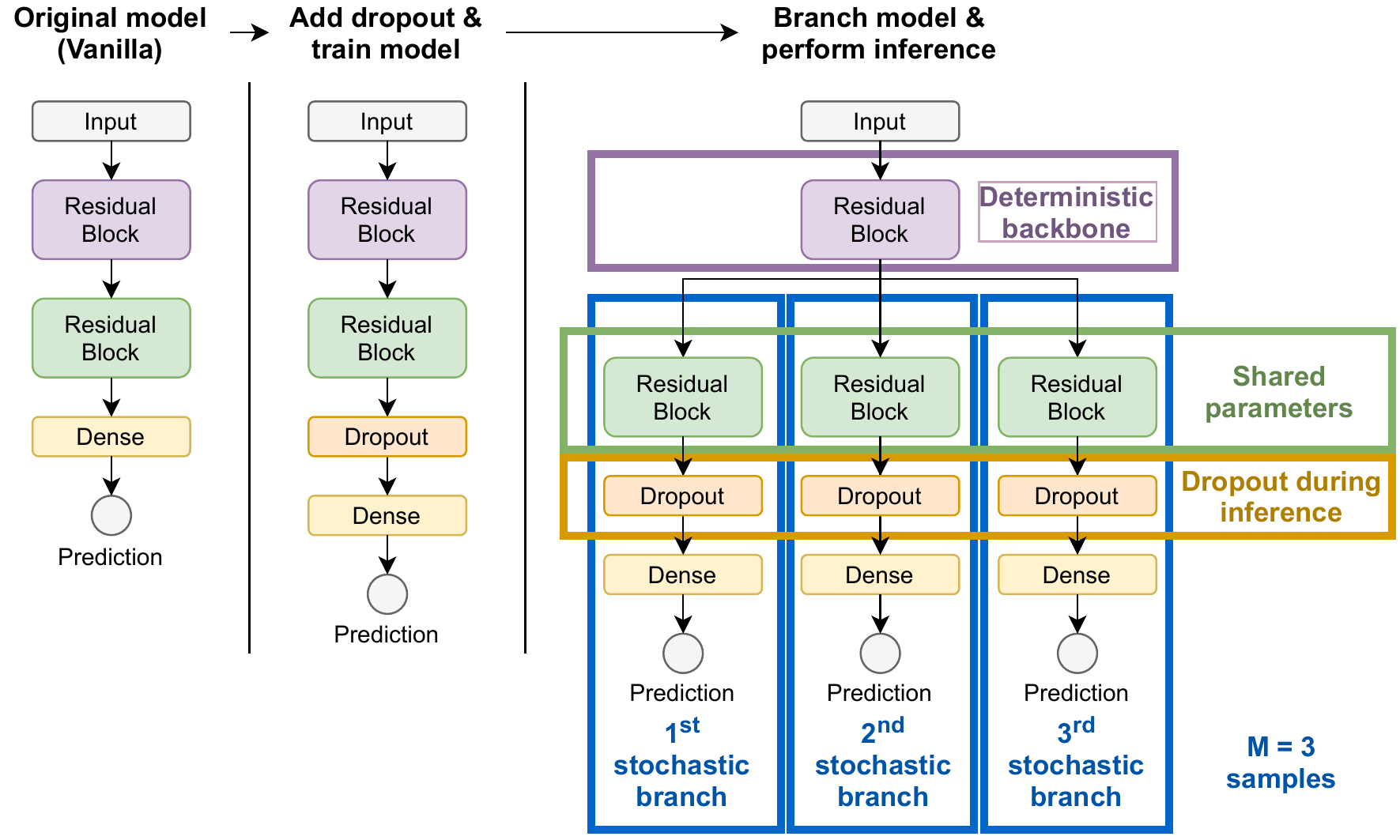}
		\caption{A high-level diagram of an Axolotl model with three branches and shared parameters. During inference, the common and deterministic backbone is computed and cached; its outputs are connected to each branch, i.e. one replica of the stochastic portion of the network, which are processed in parallel.}
		\label{fig:axolotl}
	\end{figure}

	We consider Convolutional Neural Network (CNN) based architectures and aim to (a) improve MCDO performance through diversification of sampled subnetworks, and (b) improve latency by exploitation of dropout patterns as well as by enabling batched computation on branched multi-head models. Diversification of MCDO subnetworks essentially means diversifying the considered feature subsets, which results in more diverse predictions that are, in combination, more robust~\cite{fort2019deep}. Structured dropout in CNNs enables computational optimization by fusing with convolution layers. Additionally, batched computation can be applied when using branched partial MCDO, i.e. dropout is applied only to deeper convolutional layers rather than all layers. In branched partial MCDO, as depicted in Figure~\ref{fig:axolotl}, the common backbone is computed once and cached. The subsequent stochastic model paths are computed individually and in parallel by feeding in the cached output from the backbone. Note, that all MCDO samples (subnetworks) are based on the same super network and thus share weights also in their stochastic paths.
	
	Our experiments were conducted using the Axolotl framework \footnote{\url{https://github.com/ARM-software/fast-probabilistic-models}}. This paper's main contribution is the introduction of a novel approach that (1) improves predictive performance by diversification of MCDO subnetworks using spatial dropout and contrastive dropout rates during training and inference, (2) exploits spatial convolutions and batched computation on branched models achieving low-latency inference on mobile platforms, and (3) is demonstrated to benefit uncertainty-aware human body segmentation which is a representative task for VR/AR on mobile platforms.
	
	\section{Methods}
	
\begin{figure}[th]
		\centering
		\includegraphics[width=5cm]{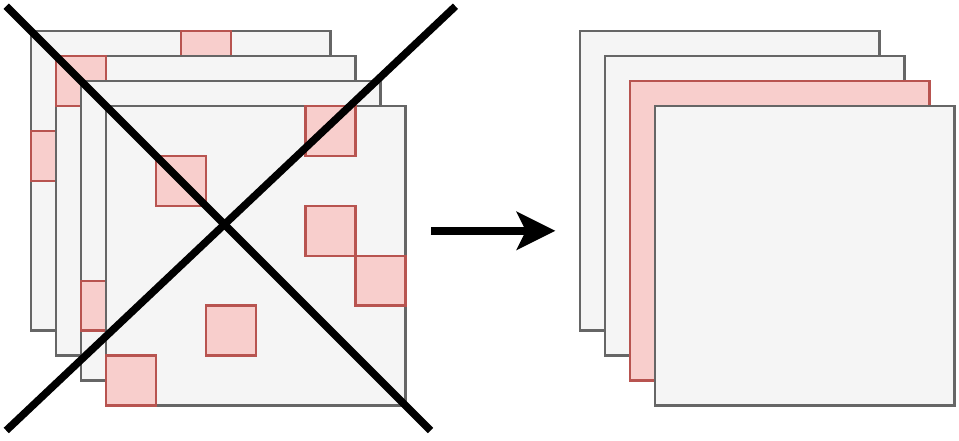}
		\caption{Spatial dropout (right) removes whole feature maps instead of randomly dropping activations (left).}
		\label{fig:do_patterns}
	\end{figure}

	\paragraph{Spatial Monte Carlo Dropout (SMCDO)}
	\label{sec:spatial_do}
	Structured dropout patterns have been used for regularization of CNNs~\cite{tompson2015efficient,notin2020improving,ghiasi2018dropblock,pham2021autodropout}. The removal of entire features~\cite{tompson2015efficient} (spatial dropout) or slices of multiple features~\cite{notin2020improving} for different members (samples/branches) of MCDO during inference results in a diversification of considered features by definition and also prevents co-adaptation during training. See Figure~\ref{fig:do_patterns} for an illustration of spatial dropout in comparison to randomly dropping activations. Spatial dropout enables optimized computation by fusing dropout and convolution operations: instead of setting feature maps to zero (standard computation), the dropped feature maps are removed completely together with the corresponding channels in the convolution kernel. The computation is performed only on the non-dropped features which reduces the number of operations and thus the latency when used with high dropout rates.
	
	\paragraph{Contrastive dropout rates during training and inference}
	Dropout during training is typically used for regularization purposes~\cite{JMLR:v15:srivastava14a}. With large dropout rates during training, there is a large regularization effect. Thus, the model focuses on the most important rather than the less informative features, and therefore may potentially replicate important features while actually preventing feature diversity. SMCDO, however, requires models that are robust towards random feature removal which can be achieved when using large dropout rates during training. Therefore, there is a trade-off between feature diversity and robustness towards random feature removal for SMCDO that directly relates to the dropout rate during training.
	
	Furthermore, higher dropout rates during inference than during training are beneficial for SMCDO predictive uncertainty, since this increases the diversity between SMCDO samples (members/subnetworks).
	
	Therefore, we propose to use contrastive dropout rates where the dropout is slightly higher during inference than it is during training. In combination with a good trade-off solution for the train-time dropout rate, contrastive dropout is beneficial for SMCDO member diversity and yields improved uncertainty estimation capabilities.
	
\begin{figure}[th]
		\centering
		\includegraphics[width=4.5cm]{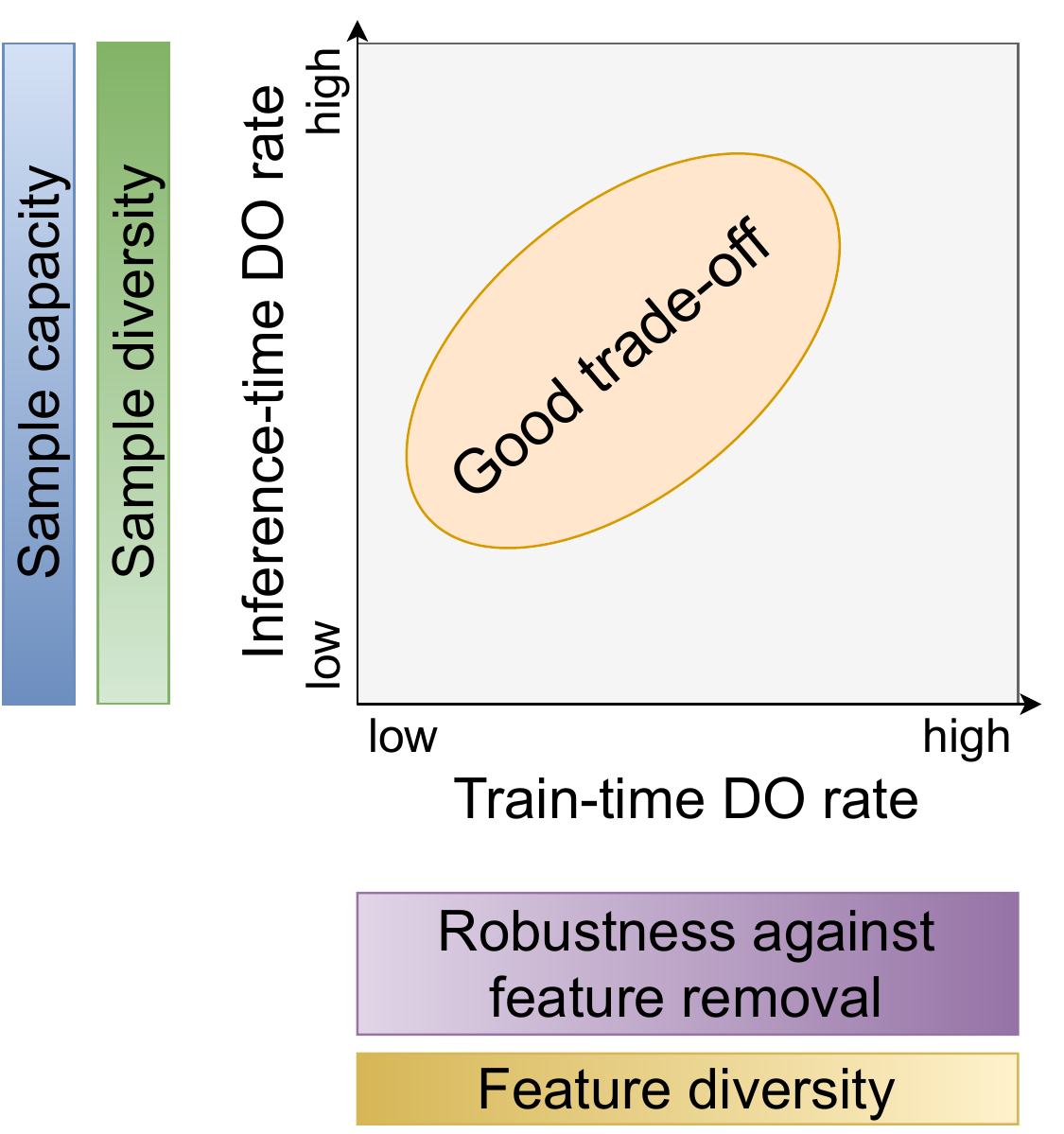}
		\caption{Illustration of the configuration space for train-time and inference-time dropout including the effects on four aspects that influence the quality of uncertainty estimates, namely sample capacity, sample diversity, robustness against feature removal, and feature diversity.}
		\label{fig:do_rates}
	\end{figure}

	The trade-off relation between train-time and inference-time dropout rates including their effects on the sample capacity, sample diversity, robustness against feature removal, and feature diversity are illustrated in Figure~\ref{fig:do_rates}.
	
	\paragraph{Speedup through Branched MCDO}
	To further improve latency we propose Branched MCDO, depicted in Figure~\ref{fig:axolotl} as a high-level diagram, where a specified number of dropout layers is added to the Vanilla model starting from the last layer (output). After training, the network is reshaped such that the deterministic backbone is shared and the subsequent stochastic portion of the network is replicated once per MCDO sample (branches). This way, the deterministic computation is performed only once and intermediate activations are cached. The branches are computed using batched inference meaning that there is only one forward pass required which improves latency considerably. Note, that Branched MCDO computation per definition does not affect predictive performance but instead reduces computational effort and thus latency.
	
	\section{Experiments}
	
	\subsection{Robust uncertainty estimation on CIFAR-10}
	Appendix~\ref{app:experiments_cifar10} details the experimental setup and Appendix~\ref{app:experiments_capacity} contains additional experiments showing that higher dropout rates for SMCDO require an increased model capacity to retain high predictive performance. Based on that, we chose Wide-ResNet20~\cite{zagoruyko2016wide} with a widening factor $k=3$ for experiments on CIFAR-10.
	
	\subsubsection{Improved predictive performance using contrastive dropout rates}
\begin{figure}[th]
		\centering
		\includegraphics[width=0.48\textwidth]{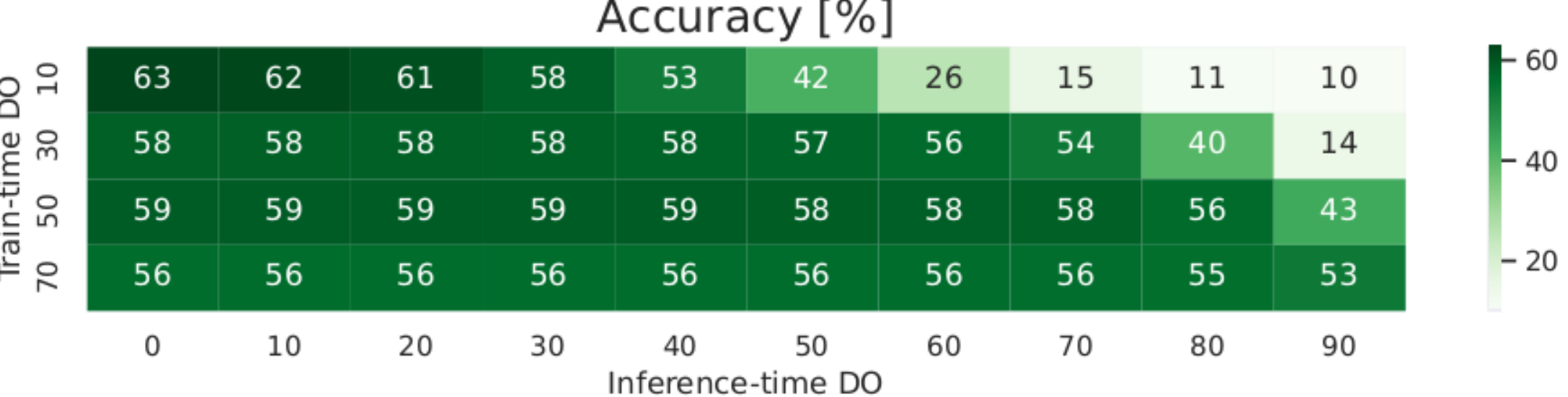}
		\includegraphics[width=0.48\textwidth]{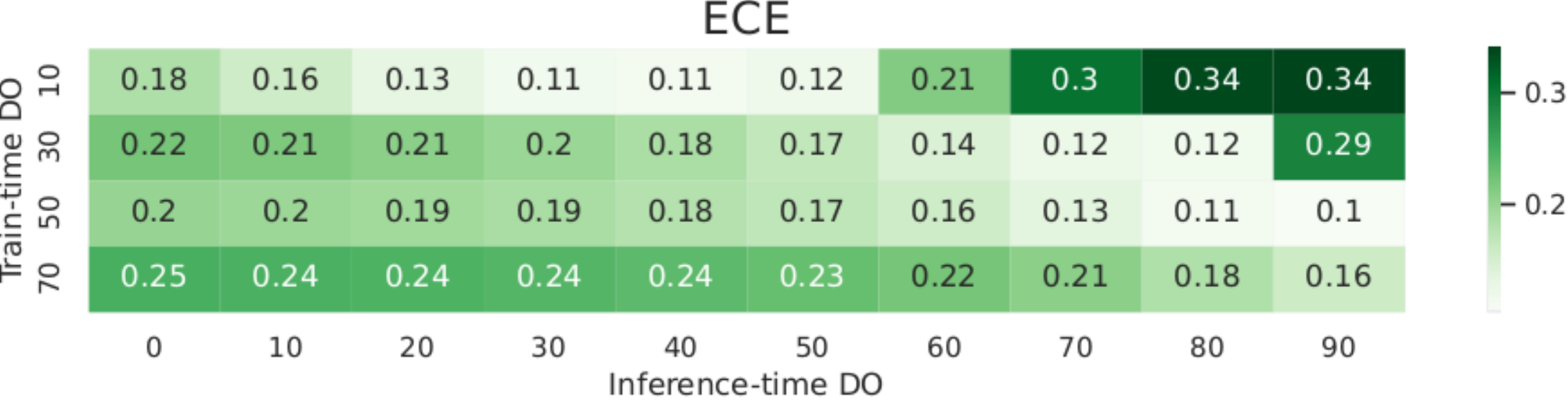}
		\caption{Accuracy and ECE for different dropout rates during training and inference. Results are computed as the mean over all corruption types for corruption level 5 (strong corruptions).}
		\label{fig:do_delta}
	\end{figure}

	Figure~\ref{fig:do_delta} shows the accuracy and expected calibration error (ECE) for different train-time dropout rates $\mathrm{DO_{train}} \in \{10\%, 30\%, 50\%, 70\%\}$ and inference-time dropout rates $\mathrm{DO_{inf}} \in \{0\%, 10\%, ..., 90\%\}$. Metrics are computed as the mean over all corruption types for corruption level 5 (strong corruptions).
	
	With larger dropout rates during training $\mathrm{DO_{train}}$, the accuracy decreases slightly. With increasing dropout rates during inference, the accuracy decreases fast with a small $\mathrm{DO_{train}}$, or it decreases slowly or stays the same with a larger $\mathrm{DO_{train}}$. Also, the ECE worsens (increases) slowly with larger $\mathrm{DO_{train}}$. However, the ECE improves (decreases) rapidly with increasing dropout rates during inference. In particular, the ECE improves considerably, when $\mathrm{DO_{inf}}$ exceeds $\mathrm{DO_{train}}$ by $10\%$ to $50\%$. Therefore we conclude, that a delta in train-time and inference-time dropout rates improves the predictive uncertainty considerably while retaining the accuracy at a high level.
	
	Generally, high dropout rates during inference improve uncertainty estimation capabilities given that the model is robust towards random feature removal. High dropout rates during training provide such a robustness towards random feature removals, however, too large dropout rates during training over-regularize model parameters and may hurt performance and feature diversity.
	
	\subsubsection{Deep Ensemble performance}
\begin{figure}[th]
		\centering
		\includegraphics[width=0.45\textwidth]{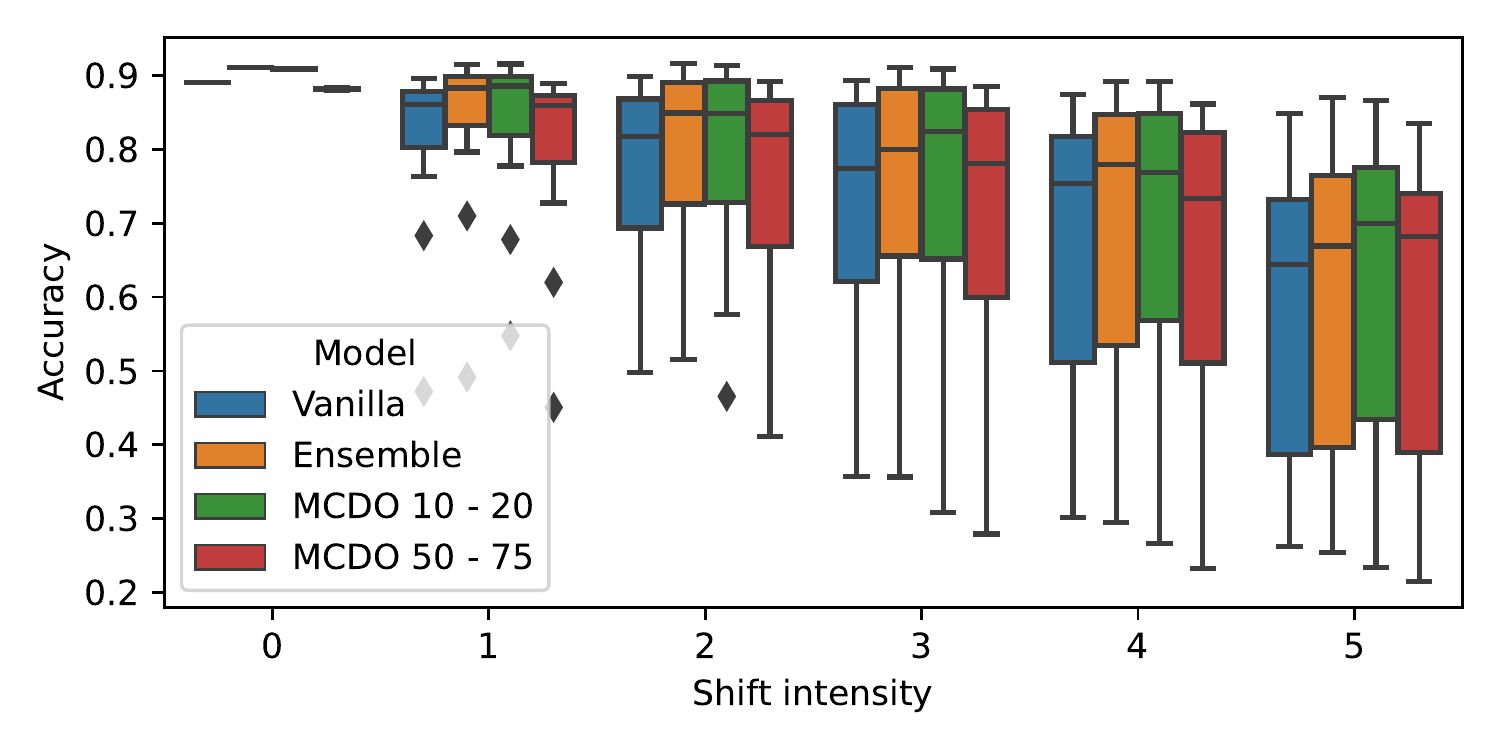}
		\includegraphics[width=0.45\textwidth]{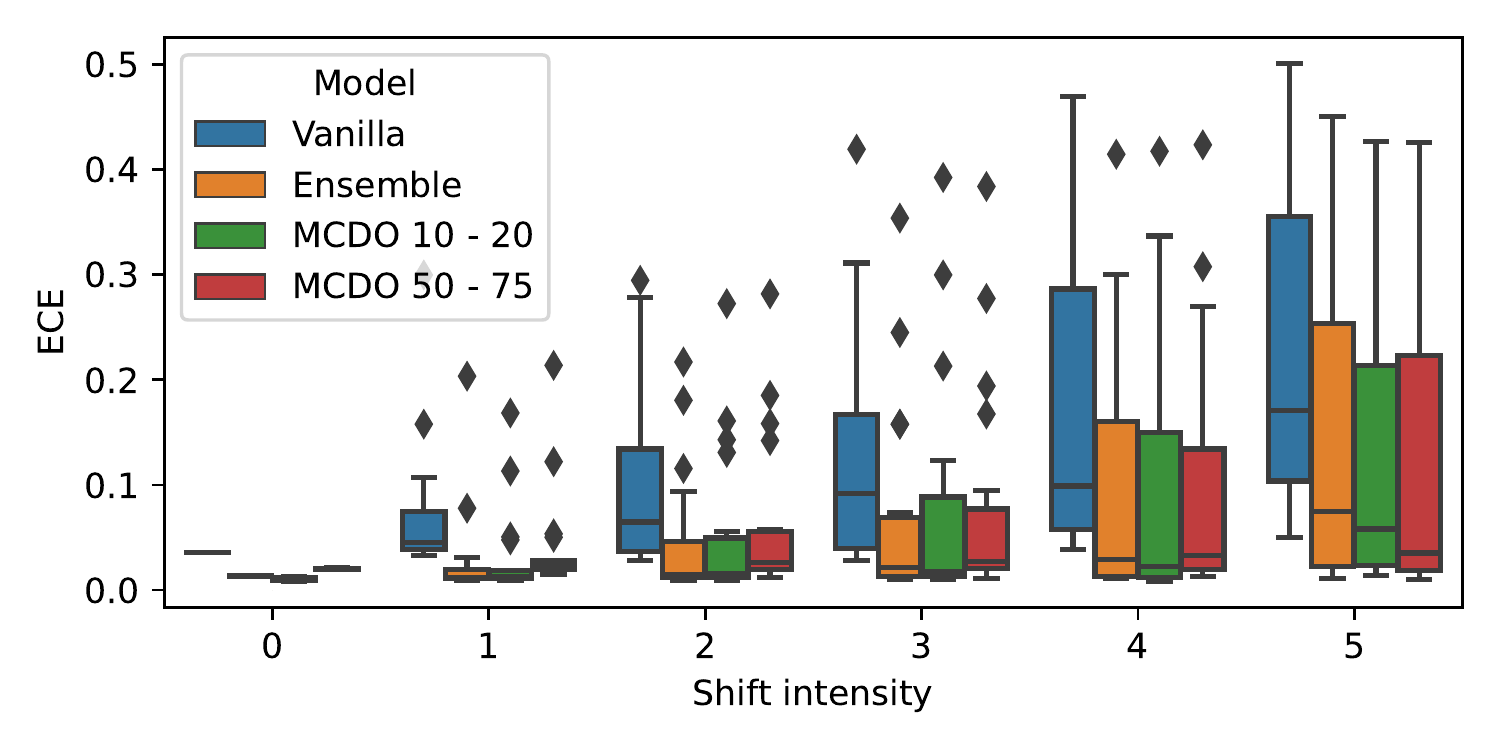}
		\caption{Accuracy and ECE comparison of Vanilla, Deep Ensemble, SMCDO with $\mathrm{DO_{train}}=10\%$ and $\mathrm{DO_{inference}}=30\%$, and SMCDO with $\mathrm{DO_{train}}=50\%$ and $\mathrm{DO_{inference}}=75\%$. X-axis shows different corruption levels and boxplot results contain metrics over all corruption types.}
		\label{fig:deep_ensemble_perf}
	\end{figure}

	Figure~\ref{fig:deep_ensemble_perf} shows accuracy and ECE for two selected configurations of SMCDO. The best performance is achieved by SMCDO with $\mathrm{DO_{train}}=10\%$ and $\mathrm{DO_{inf}}=30\%$, which has similar or better predictive performance than Deep Ensemble in terms of both accuracy and ECE. Particularly for inputs with strong corruption SMCDO yields even better results. SMCDO with $\mathrm{DO_{train}}=50\%$ and $\mathrm{DO_{inf}}=75\%$ yields slightly worse results in terms of accuracy and ECE. Note, however, that high dropout rates enable an optimized computation of fused dropout and convolution operations. For a dropout rate of 75\% our preliminary results indicate a speedup of up to $3.8 \times$ in comparison to standard computation! This configuration is therefore promising in terms of predictive performance to latency trade-off.
	
	\subsection{Uncertainty estimation in human body segmentation on mobile platforms}
	In this experiment, we use our approach, namely Branched SMCDO with contrastive dropout rates during training and inference, and demonstrate its usefulness for human body segmentation. This task (a) benefits from accurate uncertainty estimates, (b) is relevant for mobile platforms and (c) requires resource-intensive models such that efficient methods for uncertainty estimation are required and Deep Ensemble is not a viable alternative due to limited power and low-latency constraints. See Appendix~\ref{app:experiments_segmentation} for details about the experimental setup.
	
\begin{figure}[hbt]
		\centering
		\includegraphics[width=0.9\textwidth]{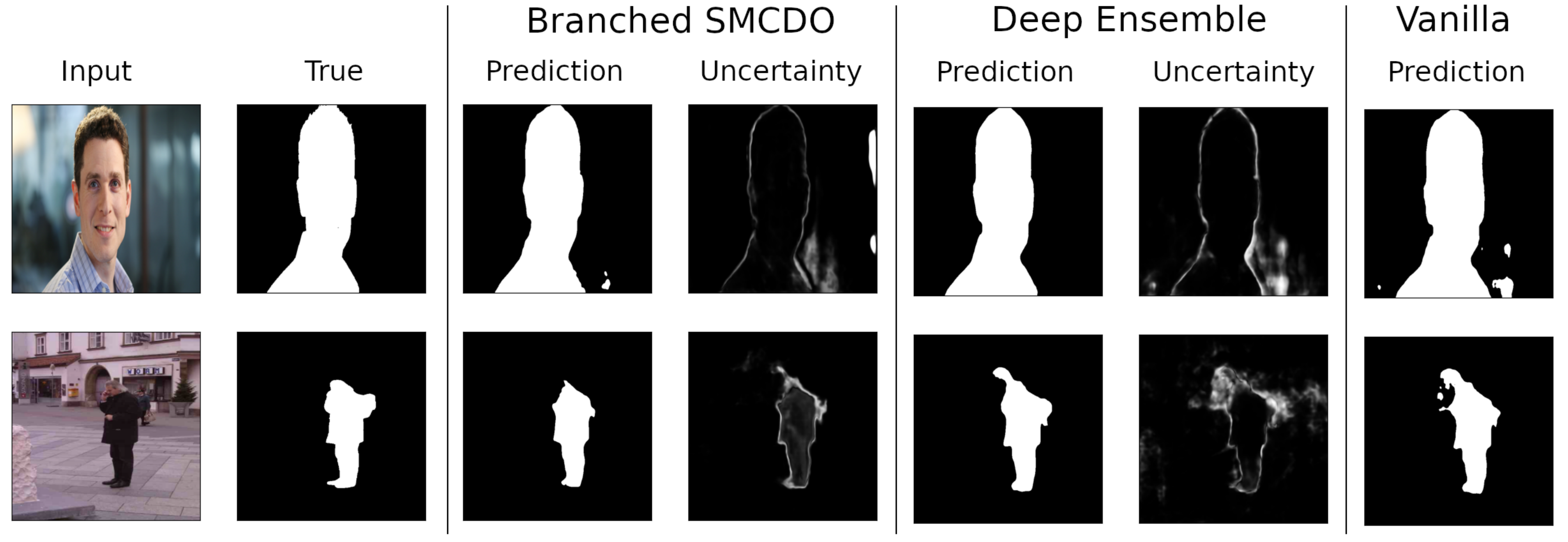}
		\caption{Exemplary segmentation predictions and uncertainties using Vanilla, Deep Ensemble and Branched SMCDO.}
		\label{fig:seg_examples}
	\end{figure}

	\paragraph{Predictive performance}
	Figure~\ref{fig:seg_performance} shows the dice score versus pixel-wise ECE for different configurations of Branched SMCDO. Vanilla (Square) and Deep Ensemble (Star) are included as reference. Circular markers represent our approach, where color encodes train-time dropout and size corresponds to inference-time dropout. Good configurations include train-time dropout of $10\%$ and $30\%$, and inference-time dropout between $10\%$ and $50\%$. The red border marks the configuration with $10\%$ train-time and $30\%$ inference-time dropout, which achieves a good trade-off between dice score and ECE. It is considerably better than Vanilla and close to Deep Ensemble predictive performance while requiring less memory and latency than the latter. Figure~\ref{fig:seg_examples} shows exemplary predictions and their per-pixel uncertainty comparing Vanilla, Deep Ensemble and our approach.
	
	\begin{minipage}[th]{\textwidth}
		\begin{minipage}{0.45\textwidth}
			\centering
			\includegraphics[height=5.2cm]{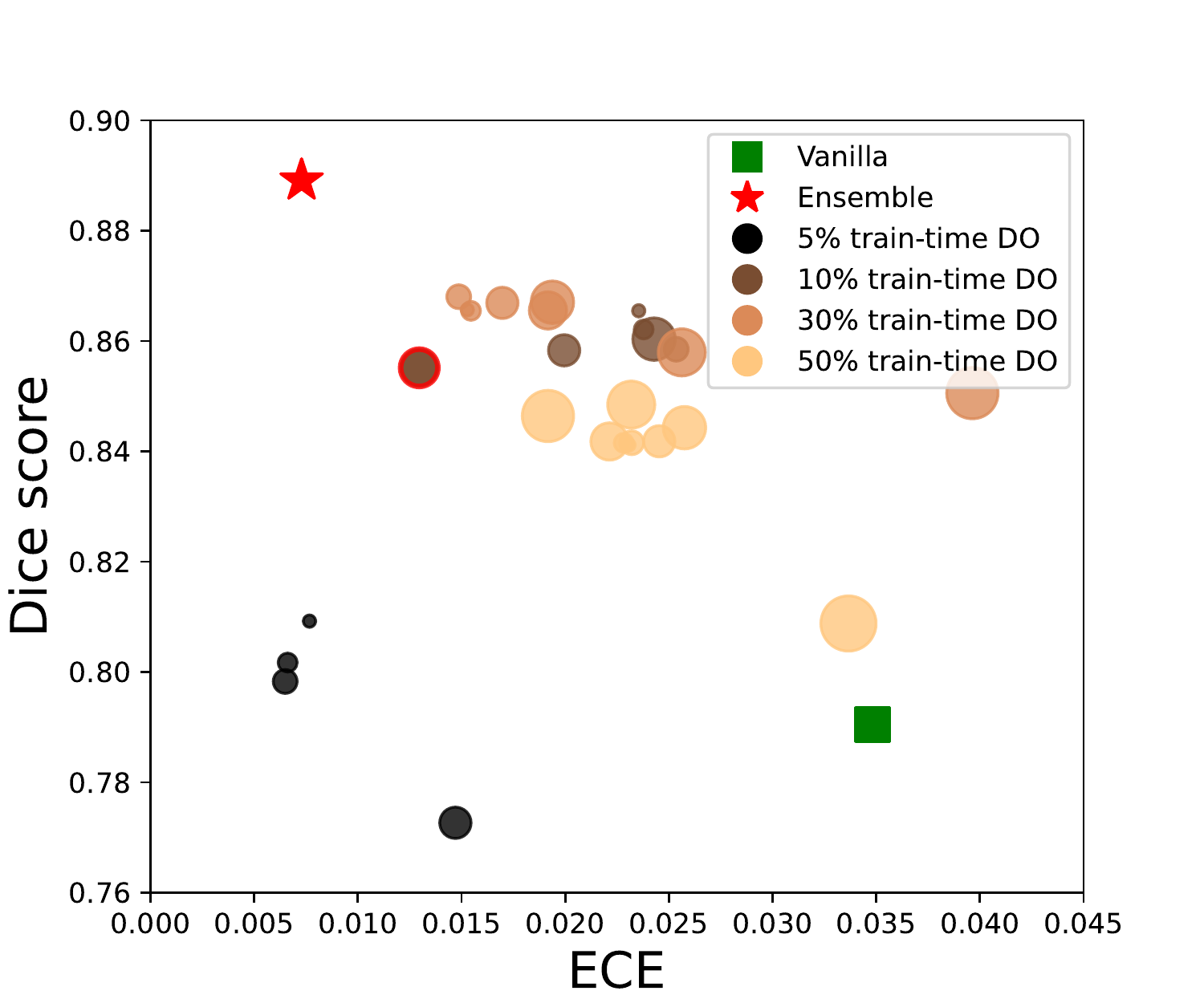}
			\captionof{figure}{Dice score versus pixel-wise ECE for human body segmentation. Circular markers represent SMCDO configurations; color and size correspond to train-time and inference-time dropout rates, respectively.}
			\label{fig:seg_performance}
		\end{minipage}
		\hfill
		\begin{minipage}{0.53\textwidth}
			\centering
			\scriptsize
			\captionof{table}{Latency in seconds and overhead relative to the Vanilla model. Measurements were taken on an NVIDIA Jetson Nano, executing on a quad-core Arm Cortex-A57 CPU.}
			\label{tab:latency}
			\begin{tabular}{lllll}
				\toprule
				Model & Latency & Overhead & Dice & ECE \\
				\midrule
				Vanilla & 0.9 & 1 & 0.7905 & 0.0348  \\
				Deep Ensemble & 2.7 & 3 & 0.8891 & 0.0073  \\
				SMCDO & 2.8 & 3.1 & 0.8551 & 0.013 \\
				Branched SMCDO & 1.4 & 1.6 & 0.8551 & 0.013 \\
				\bottomrule
			\end{tabular}
		\end{minipage}
	\end{minipage}

	\paragraph{Speedup using Branched SMCDO}
	Table~\ref{tab:latency} lists the latency of standard SMCDO in comparison with Branched SMCDO, where the shared backbone of SMCDO samples is computed only once and used as input for all branches (deterministic portion of network). Vanilla and Deep Ensemble latency is included for reference. Branched SMCDO requires only halve the latency in comparison to standard SMCDO computation or Deep Ensemble! Note, that additional speedup can be achieved by exploiting sparsity due to spatial dropout with high dropout rates as mentioned in Section~\ref{sec:spatial_do}.
	
	\section{Conclusion}
	We propose an extension of MCDO, namely (a) the use of spatial dropout, (b) contrastive dropout, and (c) branching. We show, that this combination yields improved predictive performance and enables speedup due to optimization. We obtain close to Deep Ensemble performance in terms of accuracy and ECE, but with lower latency. Experiments on the uncertainty-aware human body segmentation task show the applicability to a wider range of applications and low-latency capabilities on mobile platforms.

	\bibliographystyle{ieeetr}
	\bibliography{bibliography}
	
	\appendix
	
	\section{Experimental setup for CIFAR-10}
	\label{app:experiments_cifar10}
	
	We use the CIFAR-10 data set for training, and additionally the corrupted CIFAR-10 test set for evaluation, including 19 corruption types and 5 corruption levels. We use SGD for 200 epochs of training and standard data augmentation. We use a standard step-wise learning rate scheduler, i.e. at epochs 1, 80, 120, 160, 180 with learning rates 0.1, 0.01, 0.001, 0.0001, 0.0005. For SMCDO we use Wide-ResNet-20-k with a widening factor $\mathrm{k}=3$ and $M=3$ samples. Dropout is added to approximately the second halve of the model (Convolution layer 13 and deeper), since previous experiments have shown that this configuration yields a good performance to latency trade-off~\cite{azevedo2020stochastic}. Dropout layer are always positioned before a convolution layer. As baselines we use Vanilla ResNet-20 and Deep Ensemble with ResNet-20 and $M=3$ members. We used the Axolotl framework \footnote{\label{axolotl_fn}\url{https://github.com/ARM-software/fast-probabilistic-models}} and Keras for implementation.
	
	\section{Experimental setup for human body segmentation}
	\label{app:experiments_segmentation}
	We use a human body segmentation data set \footnote{\label{seg_fn}\url{https://github.com/VikramShenoy97/Human-Segmentation-Dataset}.} and ENet~\cite{paszke2016enet} model architecture. For training we use the Adam optimizer for 300 epochs and resize inputs to $640 \times 640$ pixels. Dropout is added to the layers from Bottleneck40 and above, where dropout is always added before a convolution layer. We use $M=3$ members for SMCDO. As baselines we use Vanilla ENet and Deep Ensemble with ENet and $M=3$ members. Models were evaluated on NVIDIA Jetson Nano executing on a quad-core Arm Cortex-A57 CPU for latency benchmarks. We used the Axolotl framework \footref{axolotl_fn} and Keras.
	%for implementation.
	
	\section{Increasing model capacity for higher dropout rates}
	\label{app:experiments_capacity}
	
\begin{figure}[hbt]
		\centering
		\includegraphics[width=4cm]{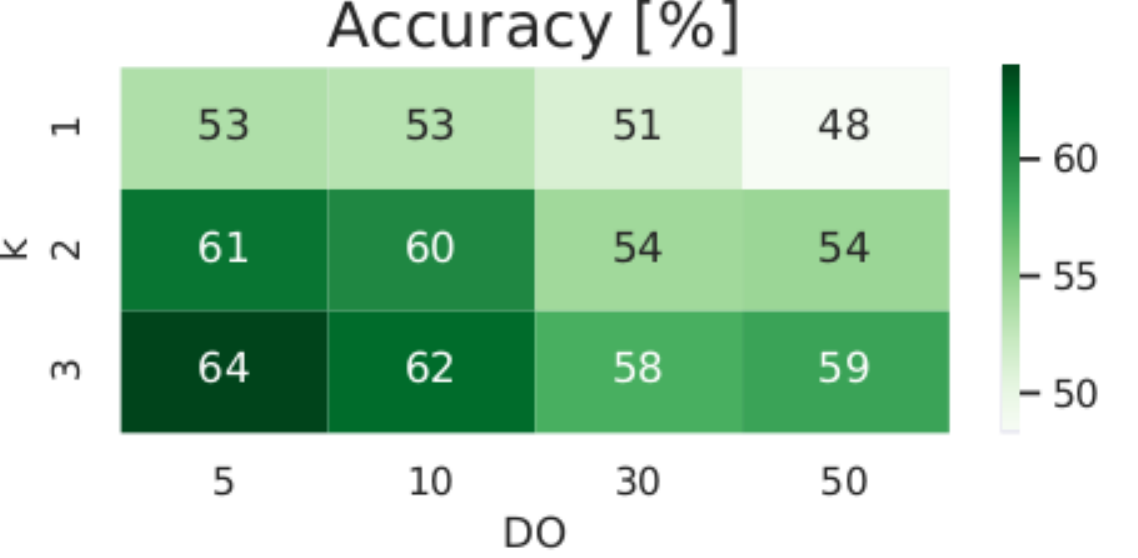}
		\includegraphics[width=4cm]{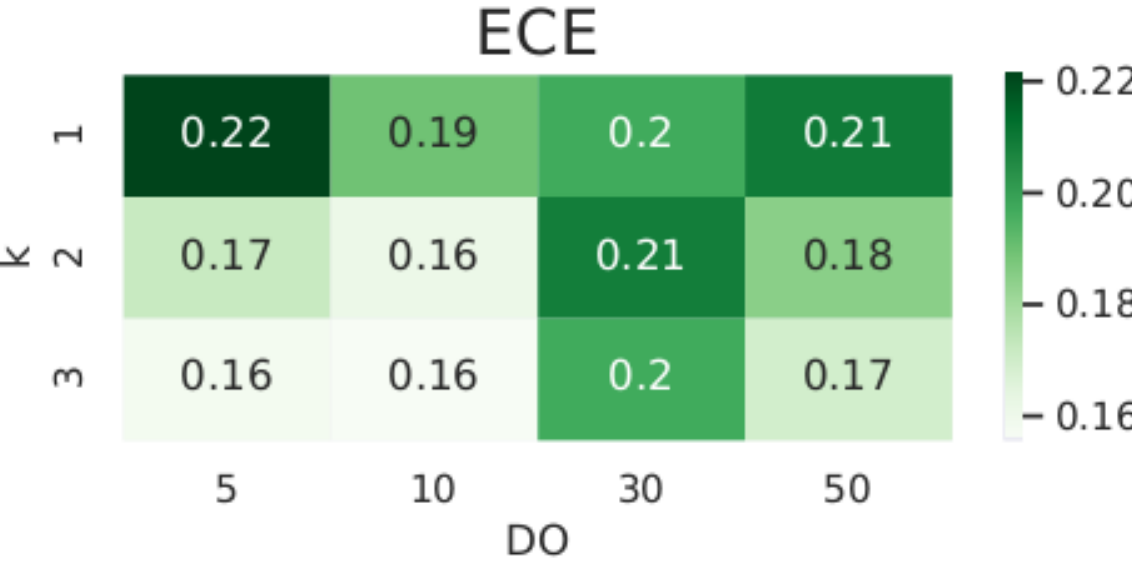}
		\caption{Accuracy and ECE for different widening factors k and dropout rates DO. Results are computed as the mean over all corruption types for corruption level 5 (strong corruptions).}
		\label{fig:k_do}
	\end{figure}

	We analyze the relation between model capacity, hence the widening factor k, the SMCDO dropout rate and the predictive performance. Figure \ref{fig:k_do} shows the accuracy and ECE for different widening factors $\mathrm{k}=\{1,2,3\}$ and dropout rates $\mathrm{DO}=\{5\%, 10\%, 30\%, 50\%\}$. Metrics are computed as the mean over all corruption types for corruption level 5 (strong corruptions). The results show that higher dropout rates require larger model capacity to retain accuracy, while slightly higher dropout rates have positive effects on the ECE (e.g. $\mathrm{DO}=10\%$). Therefore, we chose a widening factor $k=3$ for all SMCDO experiments using ResNet20 and CIFAR-10. Note that a widening factor $k=3$ increases the number of channels and thus requires more memory; the latency may or may not be affected, depending on the target device.
	
\end{document}